\documentclass[runningheads]{llncs}
\usepackage{graphicx}
\usepackage{amsmath}

\usepackage{tabularray}
\usepackage{array}
\usepackage{booktabs} % For better-looking horizontal lines
\usepackage{multirow} % For multirow cells
\usepackage{tabularx} % For setting table width

\begin{document}
\title{A Study of Age and Sex Bias in Multiple Instance Learning based Classification of Acute Myeloid Leukemia Subtypes}
\titlerunning{Age and Sex Bias in AML Subtype Classification}

\author{Ario Sadafi\inst{1,2}
\and Matthias Hehr \inst{1,3}
\and Nassir Navab \inst{2,4}
\and Carsten Marr\inst{1}
\thanks{carsten.marr@helmholtz-munich.de}}
%index{Sadafi, Ario}
%index{Hehr, Matthias}
%index{Navab, Nassir}
%index{Marr, Carsten}

%
\authorrunning{Sadafi et al.}

\institute{Institute of AI for Health, Helmholtz Zentrum München – German Research Center for Environmental Health, Neuherberg, Germany
\and
Computer Aided Medical Procedures (CAMP), Technical University of Munich, Germany
\and
Laboratory of Leukemia Diagnostics, Department of Medicine III, University Hospital, Ludwig-Maximilian University Munich, Germany
\and
Computer Aided Medical Procedures, Johns Hopkins University, USA
}
\maketitle       
\begin{abstract}
 Accurate classification of Acute Myeloid Leukemia (AML) subtypes is crucial for clinical decision-making and patient care. In this study, we investigate the potential presence of age and sex bias in AML subtype classification using Multiple Instance Learning (MIL) architectures. To that end, we train multiple MIL models using different levels of sex imbalance in the training set and excluding certain age groups. To assess the sex bias, we evaluate the performance of the models on male and female test sets. For age bias, models are tested against underrepresented age groups in the training data.
We find a significant effect of sex and age bias on the performance of the model for AML subtype classification. Specifically, we observe that females are more likely to be affected by sex imbalance dataset and certain age groups, such as patients with 72 to 86 years of age with the RUNX1::RUNX1T1 genetic subtype, are significantly affected by an age bias present in the training data. 
Ensuring inclusivity in the training data is thus essential for generating reliable and equitable outcomes in AML genetic subtype classification, ultimately benefiting diverse patient populations.
    
\end{abstract}

\section{Introduction}
Long before the advent of modern deep learning algorithms, fairness had already been a concern within the medical community. For instance, disparities in symptoms between men and women experiencing heart attacks led to differing treatment approaches \cite{ayanian1991differences}.
The machine learning solutions that the community is developing for countless medical applications is not aware of bias-related challenges that persist within medicine. Healthcare practitioners adhere to an oath that mandates unbiased treatment regardless of age, disease or disability, creed, ethnic origin, or sex \cite{parsa2017revised}. 
Fairness remains an ever-present concern in healthcare systems as computer-aided diagnosis systems play an increasingly prominent role in medical decision-making, leading to a heightened awareness of potential biases in training data. 
Biased algorithms exhibit varying performance when assessed within sub-groups defined by attributes such as sex, gender, age, ethnicity, socioeconomic status, and other related factors \cite{lara2023towards}. 
Fairness evaluation has become essential to establish the credibility of the developed systems and the availability of data and models provides an avenue for identifying and rectifying these biases.

Acute Myeloid Leukemia (AML) is a critical hematopoietic malignancy that demands accurate subtype classification for effective clinical decision-making and optimized patient care. The ability to precisely identify different AML subtypes can significantly impact treatment strategies and prognosis. For instance, acute promyelocytic leukemia, a genetic subtype of AML with a PML::RARA gene fusion is considered an oncological emergency, where rapid and appropriate therapy is crucial and curative \cite{lo2010front}.

Several automated methods are developed for single cell analysis in peripheral blood \cite{sharma2022deep}\cite{hiremath2010automated}\cite{sadafi2019multiclass}\cite{sadafi2023redtell}. Earlier works for the AML subtype classification \cite{matek2019human}\cite{salehi2022unsupervised} mostly focus on single white blood cell classification. Sidhom et al \cite{sidhom2021deep} suggest a method of combining single cell level and patient level information to provide a rapid, accurate physician-aid for diagnosing AML with PML::RARA fusion using peripheral smear images.  
In contrast, more recent methods use multiple instance learning (MIL) that enables training a model with patient-level diagnosis rather than single-cell labels. Attention-based MIL \cite{ilse2018attention} combines MIL with a trainable attention module, allowing the model to focus on specific instances within a patient's set of single-cell images for diagnosis \cite{sadafi2020attention}.
Hehr et al. \cite{hehr2023explainable} suggest an explainable MIL model to accurately classify AML genetic subtypes from blood smears with a publicly available dataset \cite{hehr2023tcia}.

Recently, various publications demonstrated the bias in medical image computing. For example Lee et al. \cite{lee2022systematic} studied the impact of race and sex bias in segmentation of cardiac magnetic resonance (MR) images. Ioanno et al. \cite{ioannou2022study} demonstrate the demographic bias in segmentation of brain MR images. Puyol-Anton et al. \cite{puyol2022fairness} investigated bias in AI-based cine cardiac MR segmentation, finding statistically significant differences in segmentation performance between racial groups. 

However, to the best of our knowledge, fairness studies on hematology datasets or multiple instance learning architectures are scarce. In this paper, we are investigating possible sex and age bias in the crucial task of AML genetic subtype classification. We use the dataset and methodology of Hehr et al. \cite{hehr2023explainable} to conduct experiments with sub-groups varying in terms of sex and age representation during the training process. We examine their impact on both male and female test sets in a sex bias study, as well as the underrepresented age groups in our age bias study.

\section{Materials and methods}

\subsection{Data}
We conducted our investigation on AML using the dataset published by Hehr et al. \cite{hehr2023explainable}. The cohort selection process involved meticulous curation of 242 samples from the Munich Leukemia Laboratory blood smear archives, with a primary focus on four distinct AML genetic subtypes and healthy controls. The blood smear images were scanned and cell detection was performed with the aid of the Metasystems Metafer software. Subsequently, a deep neural network (DNN) was employed to assign quality levels to each gallery image. The dataset consisted of a total of 101,947 single-cell images, with each patient contributing 99 to 500 images, ensuring a comprehensive representation of the disease's characteristics.

\begin{figure}[t]
\centering
\includegraphics[width=\textwidth,page=1,trim=0cm 10cm 0cm 0cm,clip]{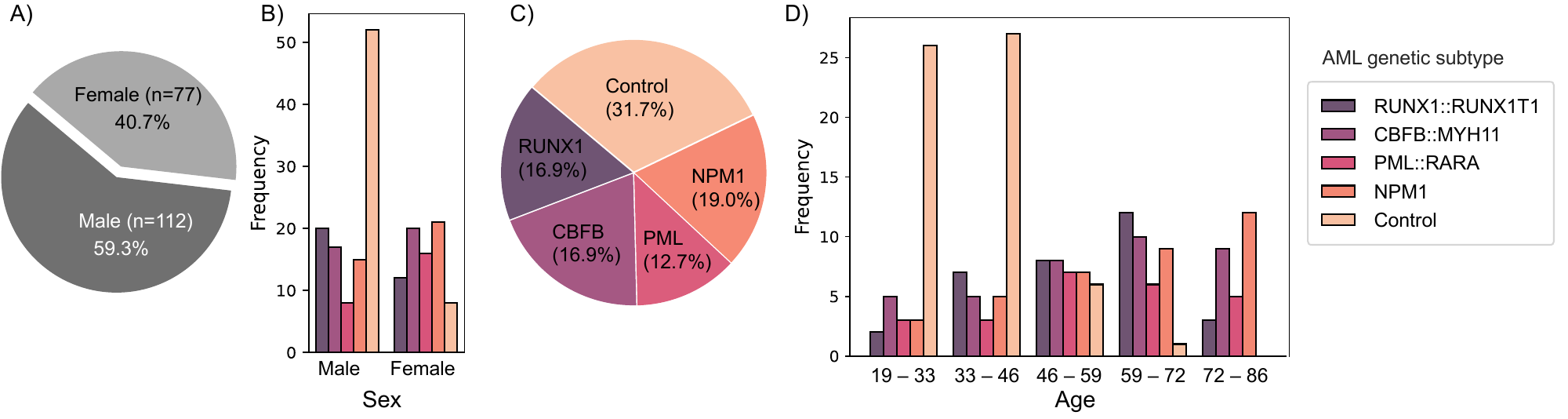}
\caption{Dataset composition. A) Distribution of sex in the dataset.
B) AML genetic subtype distribution among male and female individuals.
C) Overal distribution of AML genetic subtypes among the individuals in the dataset. 
D) Age distribution across five age groups, alongside the genetic subtype distribution within each age group.}
\label{fig:data}
\end{figure}

To ensure data quality, a comprehensive data cleaning process was implemented. This involved excluding blurry images using Canny edge detection and filtering out AML samples with a combined blast percentage of myeloblasts, promyelocytes, and myelocytes less than 20\%. Furthermore, an expert hematologist assessed sub-samples of 96 cells per patient to eliminate data artifacts. The final, filtered dataset comprised 81,214 single-cell images from 189 individuals, with each AML subtype represented as follows: APL with PML::RARA (n = 24), AML with NPM1 mutation (n = 36), AML with CBFB::MYH11 fusion (n = 37), and AML with RUNX1::RUNX1T1 fusion (n = 32), and a healthy control
group of 60 stem cell donors. Every patient sample is a single training data point and consists of 500 single cell images. Age and sex information of the patients in the dataset is also reported. Figure \ref{fig:data} provides an overview of the dataset details.

\subsection{Multiple instance learning}
The proposed approach consists of two steps: (i) A feature extractor previously trained on a similar task is used to extract features from every single cell image. (ii) Using the extracted features, an attention based MIL model is used for AML subtype classification. We briefly describe the method in the following:

\subsubsection{Feature extraction.}
With a dataset of over 300,000 single-cell images from 2205 blood smears, a ResNet34 \cite{he2016deep} model is trained to extract single-cell features. The dataset was annotated into 23 single-cell classes by experienced cytologists, excluding two classes due to their small size. Data augmentation techniques such as flipping, rotation, and random erasing were applied to the single-cell images, along with probabilistic oversampling to address class imbalance.
The model was initialized with weights trained on ImageNet \cite{russakovsky2015imagenet} and optimized using categorical cross-entropy. With the trained model a feature vector of 12,800 features is extracted from the 34th layer of the model before final pooling and fully connected layers. This feature vector represented the characteristics of the input image.
Training continued until no significant improvement in validation loss was observed for 10 consecutive epochs. The best-performing model on the test set from the cross-validation folds was chosen for feature extraction on a separately scanned multiple instance learning dataset.

\subsubsection{Multiple instance learning.}
A permutation invariant method is used to analyze a set of single-cell images from a patient and predict the associated AML subtype \cite{hehr2023explainable}. The method returns attention scores, indicating the importance of each cell in the bag for classifying different AML subtypes. The attention based MIL model works by computing embedded feature vectors from the initial feature vectors of the single-cell images. These vectors are then used to predict the patient's AML subtype based on the calculated attention scores.The class-wise attention matrix is designed to ensure that each attention value directly influences the prediction of its corresponding class, eliminating interclass competition of attention values. This way, the model can make more accurate predictions for each AML subtype.

The training procedure was identical to that of \cite{hehr2023explainable}, resulting in similar metrics as reported. Training employed a learning rate of $5\times10-5$, and stopped if no improvement in validation loss was observed for 20 epochs, with a maximum of 150 epochs.

\section{Experiments}
In this section we describe the experiments we performed to investigate possible age and sex bias of the MIL model for AML subtype classification. 

\begin{table}
  \centering
  \caption{Training and test sets for sex bias experiments. The table presents the number and percentages of male and female individuals in the entire training set. The mean and standard deviation of individuals' age is reported for each experiment. The test sets are comprised of 20\% of the total respective female and male populations.}
  \label{tab:sex}
  \begin{tabularx}{\textwidth}{llllll} % Set table width to 100% of text width
    \toprule
    Section & Female (\%) & Male (\%) & Average age $\quad$ & Description & n\\
    \midrule
    \multirow{5}{*}{Training set} & 0 & 89 (100\%) & 46.8 ± 17.5 & 0\% female & 89 \\
    & 30 (25.2\%) & 89 (74.8\%) & 49.1 ± 18.2 &  25\% female & 119 \\
    & 61 (50\%) & 61 (50\%) & 52.3 ± 18.1 &  50\% female & 122 \\
    & 61 (75.3\%) & 20 (24.6\%) & 54.9 ± 17.6 & 75\% female & 81 \\
    & 61 (100\%) & 0 & 56.7 ± 17.4 & 100\% female & 61 \\
    \midrule
    Test set (Male) & - & 23 (20\%) & 46.5 ± 20.6 & All male testset \\
    Test set (Female) & 16 (20\%) & - & 52.7 ± 17.5 & All female testset \\
    \bottomrule
  \end{tabularx}
  
\end{table}

\subsection{Sex bias}
From the total of 189 individuals, 77 are females and 112 are males. We designed 5 experiments to train a MIL model with different distributions of sex in the dataset and evaluate all of them on two hold out test sets of only male and only female individuals ensuring a stratified split to have all of the genetic subtypes present. Table \ref{tab:sex} shows the sex proportions in different experiments both in training and test splits. A 10\% validation set was sampled later from the training split. Holdout test sets are 20\% of all of the data. 

\subsection{Age bias}

To assess age bias, patients were divided into 5 distinct age groups (see Fig \ref{fig:data}c). Each experiment involved holding out one of the age groups as the test set while training the model on all other age groups. Table \ref{tab:age} displays the age groups along with the sex distribution within each group.

\begin{table}
  \centering
  \caption{Different age groups used in age bias experiments. The dataset is divided into 5 different age groups. Number of male and females in every group and percentage of males is reported for every age group.}
  \label{tab:age}
  \begin{tabularx}{0.60\textwidth}{lXc} % Set table width to 100% of text width
    \toprule
    Section $\qquad$ & Age & Male - female (\%)  \\
    \midrule
      1. Young adults & [19.8, 33.1)& 28 - 11 (72\%)\\
      2. Middle adulthood & [33.1, 46.3) & 37 - 10 (79\%)\\
      3. Midlife & [46.3, 59.5) & 16 - 20 (44\%)\\
      4. Senior adults & [59.5, 72.8) & 16 - 22 (42\%)\\
      5. Elderly & [72.8, 86.1] & 15 - 14 (51\%)\\
    \bottomrule
  \end{tabularx}
  
\end{table}

\section{Results}
We use the area under the precision-recall curve to evaluate the performance of the models for every class. Every experiment is repeated 5 times with different seeds and the standard deviation is reported. Statistical testing was performed using Kruskal-Wallis test \cite{kruskal1952use} due to its suitability for comparing independent groups with non-normally distributed data. Subsequently, we performed post hoc Dunn test \cite{dunn1964multiple} with Bonferroni correction \cite{bonferroni1936teoria} to identify significantly diverse groups.

\subsection{Sex bias}

\begin{figure}[t]
\centering
\includegraphics[width=\textwidth,page=2,trim=0cm 7.5cm 0cm 0cm,clip]{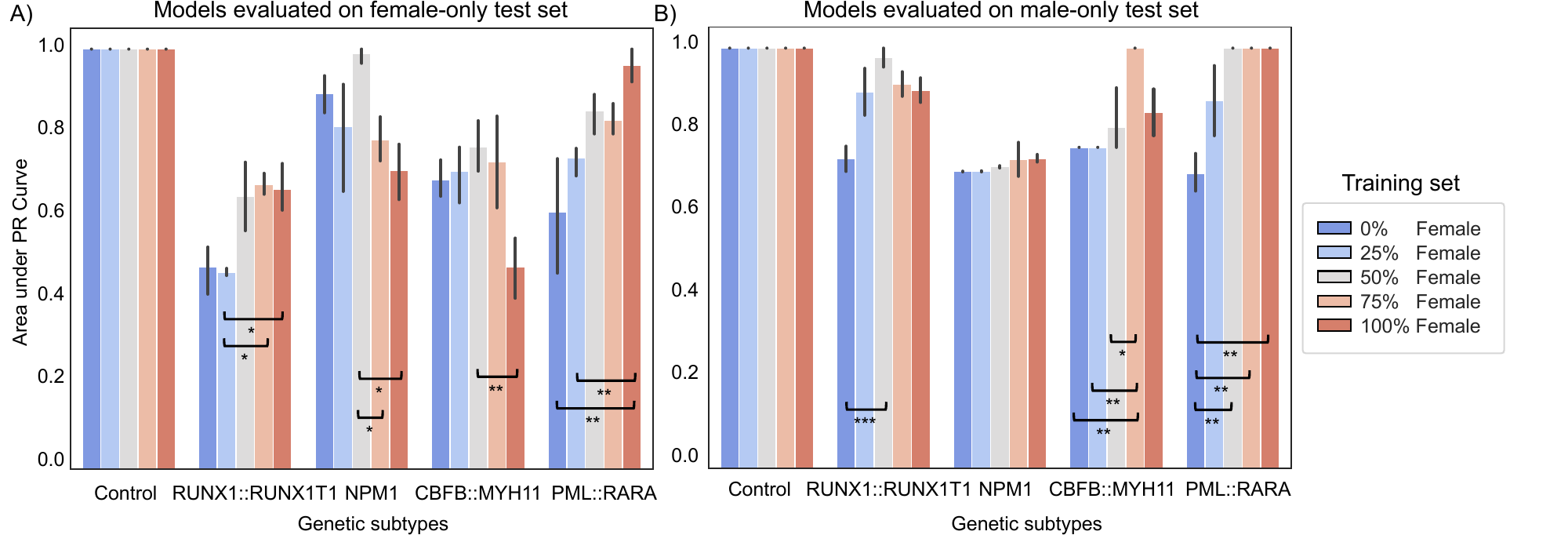}
\caption{Area under precision-recall (PR) curve for each class, representing models trained with varying proportions of females versus males in the training set. The models were tested on two separate test sets: A) the female-only test set, and B) the male-only test set. Statistical significance was determined using The statistical analysis involved the Kruskal-Wallis test followed by post hoc pairwise Dunn test with Bonferroni correction, indicated by asterisks:  * ($0.1\geq p>0.05$), ** ($0.05\geq p >0.01$), *** ($0.01\geq p >0.001$) }
\label{fig:sex}
\end{figure}

Figure \ref{fig:sex}A illustrates how all models perform on the holdout female test set under various training set conditions. Interestingly, we observed that a larger representation of females in the training data positively influences the classification of the PML::RARA genetic subtype. In Figure \ref{fig:sex}B, we examined the performance of the same models on the male test set, and once again, we noticed that a higher proportion of females in the training set contributes to improved classification of PML::RARA samples. This is evident as the area under the precision-recall curve for both genetic subtypes is significantly higher when female patients are present. Statistical tests also support these observations.

For the RUNX1::RUNX1T1 genetic subtype, the best results are obtained when a balanced proportion of males and females are present in the training set. Conversely, when only male patients are present in the training set, a significantly lower performance is observed in the evaluation on both test sets.

In the female test set, NPM1 and CBFB::MYH11 subtypes show significantly better performance with a balanced training set, whereas in the male test set, NPM1 does not seem to depend much on the patients' sex in the training set. On the other hand, CBFB::MYH11 shows significantly better performance with more female representation in the training set.

Across both the male and female test sets, the models consistently achieve accurate classifications for control samples from healthy stem cell donors.

The presence of female patients in the training set has a more significant impact on the models' performance on the female test set compared to the male test set. While the male test set is generally affected to a lower degree, in most cases, a higher ratio of female to male presence in the training set positively affects the performance.

% \begin{table}[htbp]
% \centering
% \caption{Accuracy and fairness metrics reported for the sex bias experiments }
% \label{tab:results}
% \begin{tabular}{ccccccc}
% \toprule
% Train on & Test on & Accuracy & F1 score & Eopp1 & Eodd & Theil index \\
% \midrule
% Exp0 & Male & 0.90 $\pm$ 0.02 &  0.82 $\pm$ 0.03 & -0.00168 ± 0.03676 & & \\
%  & Female &0.49 $\pm$ 0.07 &  0.50 $\pm$ 0.10& & & \\
% Exp1 & Male & 0.90 $\pm$ 0.03 &  0.82 $\pm$ 0.04 & &  & \\
%  & Female & 0.60 $\pm$ 0.06 &0.61 $\pm$ 0.06 & & & \\
% Exp2 & Male & 0.94 $\pm$ 0.02 & 0.91 $\pm$ 0.03 & & & \\
%  & Female & 0.68 $\pm$ 0.02 &  0.70 $\pm$ 0.02 & & & \\
% Exp3 & Male & 0.87 $\pm$ 0.03 & 0.66 $\pm$ 0.03  &  && \\
%  & Female & 0.79 $\pm$ 0.05 & 0.69 $\pm$ 0.03 & & & \\
% Exp4 & Male & 0.85 $\pm$ 0.04 & 0.80 $\pm$ 0.05 & & & \\
%  & Female & 0.62 $\pm$ 0.06  & 0.65 $\pm$ 0.05 & & & \\
% \bottomrule
% \end{tabular}
% \end{table}

\subsection{Age bias}
\begin{figure}[t]
\centering
\includegraphics[width=0.80\textwidth,page=3,trim=0cm 7.3cm 11.5cm 0cm,clip]{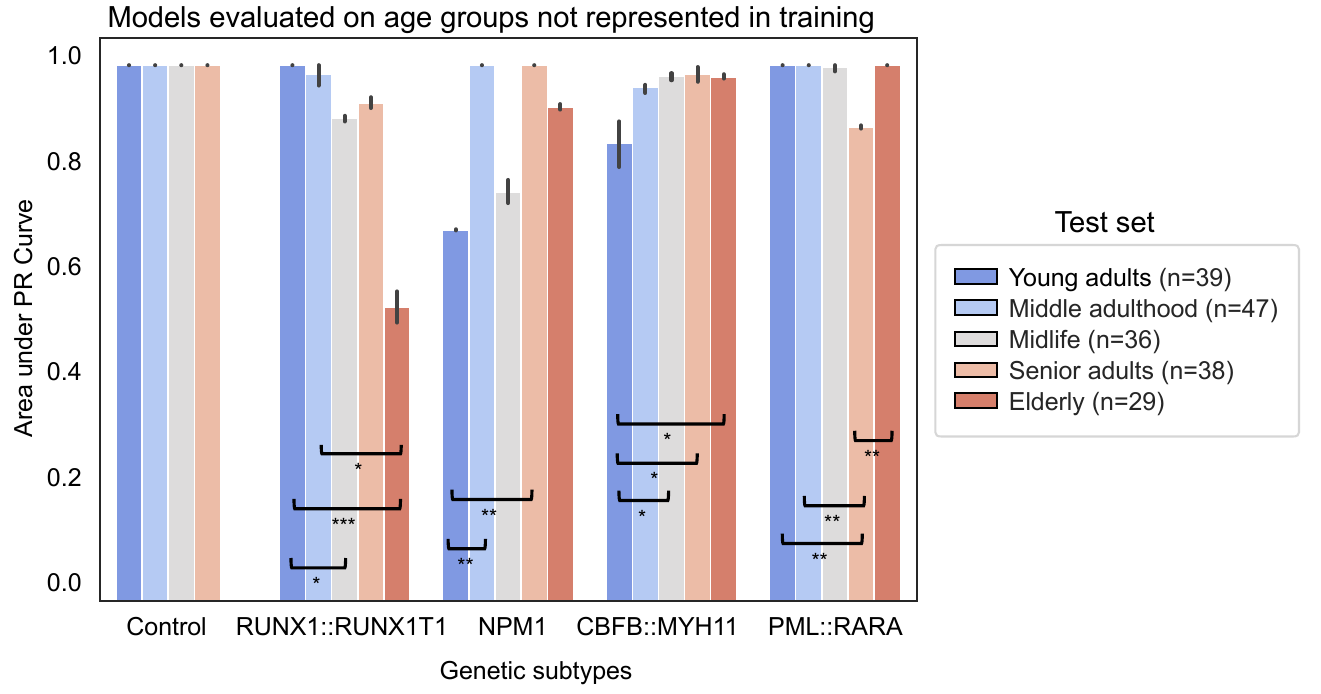}
\caption{Area under precision-recall (PR) curve for every class is reported for models trained with different age groups represented in the training set. The test set age group is always held out. Statistical significance was determined using Kruskal-Wallis test followed by post hoc pairwise Dunn test with Bonferroni correction, indicated by asterisks: * ($0.1\geq p>0.05$), ** ($0.05\geq p >0.01$), *** ($0.01\geq p >0.001$)}
\label{fig:age}
\end{figure}

For the age bias experiments, certain age groups are excluded from the training set and the performance of the models on the excluded age groups is shown in Figure \ref{fig:age}, which illustrates the area under the precision-recall curve for the models in the experiment.
% \sloppy

We observed a significant impact on elderly patients with the RUNX1::RUNX1T1 subtype in this experiment, suggesting a potential age bias against this specific age group.
The model performance in this age group was significantly lower compared to other age groups specially the young adults. This is indicating possible age bias against elderly when it comes to detecting the RUNX1::RUNX1T1 subtype of AML. 

Additionally, the absence of young adults with NPM1 and CBFB::MYH11 subtypes in the training set had a significant effect on the models' performance. In both cases, the classifiers' ability to accurately classify NPM1 and CBFB::MYH11 subtypes was compromised. 

Since no elderly individuals were represented in the control samples of the dataset (Fig. \ref{fig:data}c), their performance is not reported in Figure \ref{fig:age}.

\section{Discussion}
We have demonstrated the possibility of age and sex bias in AML subtype classification with MIL architectures. Our results suggest that the absence or presence of certain age groups and sexes in the training data can significantly impact the classification performance of specific AML subtypes. In elderly, diagnosis of RUNX1::RUNX1T1 is more challenging if it is underrepresented in training. Sample quality is also worse in elderly. Interestingly, RUNX1::RUNX1T1 genetic subtype is also less common among patients (see Fig \ref{fig:data}).

Moreover, we have pointed out the possible influence of sex bias in the classification process. The variation in performance between male and female test sets indicates that certain AML subtypes may be classified more accurately when trained on balanced dataset representing both sexes. Specially, females are seen to be affected by a sex imbalanced training set more than males. This raises concerns about potential disparities in diagnosis and treatment recommendations based on sex. 
% Finally, statistical tests indicate a more pronounced variability in the experiments involving age bias compared to sex bias. 

Although MIL methods are providing explainability \cite{hehr2023explainable}\cite{sadafi2023pixel} for better clinical applicability, addressing and mitigating biases in AML subtype classification models is important for both the machine learning community and medical researchers. There is a pressing need to collect more diverse and well-balanced multicentric datasets, encompassing various age groups, sex, and ethnicities, in order to enhance the models' generalizability and fairness. Additionally, incorporating fairness-aware learning techniques and conducting comprehensive bias analysis during model development can play a significant role in reducing biases.

Naturally, the data distribution is different between our experiments as a result of sampling and this needs to be considered for the interpretation of our results. It also raises questions about fair data acquisition: Data bias does not start upon image acquisition, but needs to be considered during patient selection and study design. In the medical domain, almost every disease is distributed unevenly alongside various dependent variables such as age, sex, ethnicity or social status. Thus, methodological solutions must become part of every machine learning scientists' repertoire to cater the oath of physicians in pursuit of fair and just medical care.

\section*{Acknowledgments}
C.M. has received funding from the European Research Council (ERC) under the European Union’s Horizon 2020 research and innovation programme (Grant agreement No. 866411).

\bibliographystyle{splncs04}
\bibliography{article}
\end{document}